\documentclass{ieeeaccess}
\pdfoutput=1
\usepackage{cite}
\usepackage{amsmath,amssymb,amsfonts}
\usepackage{algorithmic}
\usepackage{graphicx}
\usepackage{textcomp}
\usepackage{array}
\usepackage[table]{xcolor}
\def\BibTeX{{\rm B\kern-.05em{\sc i\kern-.025em b}\kern-.08em
    T\kern-.1667em\lower.7ex\hbox{E}\kern-.125emX}}
\begin{document}
\doi{10.1109/ACCESS.2017.DOI}

\title{Neural Network Generalization: The impact of camera parameters}
\author{\uppercase{Zhenyi Liu}\authorrefmark{1,2},
\uppercase{Trisha Lian}\authorrefmark{1},
\uppercase{Joyce Farrell\authorrefmark{1} and Brian Wandell}\authorrefmark{1}}
\address[1]{Stanford University (e-mail: {zhenyiliu,  tlian, jefarrel, wandell}@stanford.edu)}
\address[2]{State Key Laboratory of Automotive Simulation and Control, Jilin University}
\tfootnote{Supported by Jilin University. We thank Henryk Blasinski for his contributions.}

\markboth
{Z.Liu \headeretal: Neural Network Generalization}
{Z.Liu \headeretal: Neural Network Generalization}

\corresp{Corresponding author: Zhenyi Liu (e-mail: zhenyiliu27@gmail.com)}

\begin{abstract}
We quantify the generalization of a convolutional neural network (CNN) trained to identify cars. First, we perform a series of experiments to train the network using one image dataset - either synthetic or from a camera - and then test on a different image dataset.  We show that generalization between images obtained with different cameras is roughly the same as generalization between images from a camera and ray-traced multispectral synthetic images. Second, we  use ISETAuto, a soft prototyping tool that creates ray-traced multispectral simulations of camera images, to simulate sensor images with a range of pixel sizes, color filters, acquisition and post-acquisition processing. These experiments reveal how variations in specific camera parameters and image processing operations impact CNN generalization. We find that (a) pixel size impacts generalization, (b) demosaicking substantially impacts performance and generalization for shallow (8-bit) bit-depths but not deeper ones (10-bit), and (c) the network performs well using raw (not demosaicked) sensor data for 10-bit pixels.
\end{abstract}


\titlepgskip=-15pt

\maketitle

\section{Introduction}
\label{sec:introduction}
\PARstart{T}{raining} a convolutional neural network (CNN) to detect cars requires a large amount of labeled data. To support this application, a number of groups have taken an empirical approach, acquiring data with a specific camera or a small collection of cameras that were designed for consumer photography. These data sets are then labeled, either by people alone or with the help of specialized algorithms.  Several of these datasets have been shared and serve as a resource for training \cite{Geiger2013-zh,Cordts2016-sz,Neuhold2017-pu,Yu2018-wr} 

Such an empirical approach to network training is not practical for camera design. The cost of designing and building a camera, acquiring training data, labeling the training data, and assessing the network performance is simply too expensive. Simulation of image data from the novel camera design is the only option. For this reason we are developing software tools (ISETAuto) \cite{Blasinski2018-se,Liu2019-lc,Liu2019-ji}) for simulating physically accurate, complex and realistic driving scenes, modeling camera optics and sensors, and controlling the neural networks that detect and localize objects. 

In this paper, we first address the broad question of how well training with synthetic data sets, generated by modern soft prototyping software, generalizes to camera data. Establishing the level of generalization across data sets is a central question in many neural network applications; establishing the accuracy of generalization is essential for this application.  We then show how the soft prototyping system can be used to explore how training generalizes as specific camera parameters and image processing operations are varied.

This work contributes the following:
\par 1. We quantify generalization by measuring how well a network trained on data from one imaging system performs on an independent data obtained with a second imaging system. These measurements reveal significant limits on generalization between camera images, extending earlier reports \cite{Torralba2011-pd}. 
\par 2. We analyze generalization for networks trained and evaluated on combinations of synthetic images and camera images. The generalization limitations between synthetic and camera are similar to those found between camera data sets.
\par 3. We measure generalization using images created with a soft prototyping tool that controls camera parameters (e.g., pixel size, quantization) and image processing algorithms (e.g., exposure control).  These controlled measurements quantify the impact that camera parameters have on network generalization and performance.

\section{Related work}
\label{sec:related work}
\subsection{Generalization}
How well a system generalizes between different datasets (train on A, test on B) is a key measure of computer vision object detection algorithms. Generalization was identified as a critical shortcoming of support vector machine (SVM) algorithms \cite{Torralba2011-pd}.  The size and complexity of image datasets has increased significantly over the past decade. In the case of automotive applications the development of region proposal networks (RPN) has extended the applications from databases containing images of single objects (e.g., CalTech 101 and ImageNet) to images representing more complex scenes (BDD, KITTI, Cityscape). Quantifying generalization of neural networks between these complex image datasets remains an important question \cite{Movshovitz-Attias2016-bg}. Understanding generalization can not only help us understanding the limit of the deep neural network, but also guide us feeding more suitable data to it.

\subsection{Synthetic data methods}
Because of the challenges in obtaining large empirical datasets, there has been much interest in using computer graphics to generate synthetic datasets to use for training. A number of authors consider specific methods of creating synthetic images for automotive applications, and they draw different conclusions about the ability of these methods to generalize to camera images \cite{Vazquez2014-vp,Sun2014-mx,Ros2016-ub,Movshovitz-Attias2016-bg,Tremblay2018-az,Wrenninge2018-ri,Carlson2019-oe}.  These papers include useful ideas about how to create images that improve generalization between synthetic and camera datasets. 

One method uses computer graphics to augment camera images, superimposing virtual objects of the target category onto the image \cite{Alhaija2017-kz,Li2019-gc}.  This approach is simpler than synthesizing entire scenes and improves generalization.  A second method is to build scenes and objects using game engine technology \cite{Kar2019-zn,Tremblay2018-hs}. This approach is helpful for building scenes; its limitation concerns the image quality of rendering. Third, some authors use ray-tracing to produce high quality images \cite{Wrenninge2018-ri,Blasinski2018-se}. Ray-tracing has been combined with driving simulators to automate the creation of large numbers of realistic scenes \cite{Liu2019-lc}. A limitation of ray-tracing is that the rendering time is long compared to game engines, real-time ray-tracing is a current undergoing research topic \cite{Haines2019-pb}, which can potentially be helpful. 
\subsection{Synthetic data motivations} 
It is worth distinguishing three purposes for using synthetic data.  One is to build a single network that performs well on data from any camera. It is not yet known what the limits of network performance might be; the human ability to recognize objects across many types of cameras supports the idea that a single network might be able to succeed with images from a wide range of cameras \cite{Geirhos2018-yg}.  Massive simulation, say by generating a very large number of images from many different cameras with diverse scene content, is one approach to creating such a universal network \cite{Ros2016-ub,Tremblay2018-hs,Wrenninge2018-ri,Prakash2018-ep}. This approach synthesizes images as a data augmentation method.

A second purpose is to improve a network’s performance on images from a particular camera.  A typical approach is to augment labeled data from the specific camera with synthetic data. A number of different methods are available to make synthetic images match existing camera images. For example, style transfer has been used \cite{Carlson2019-oe} and networks have been proposed to generate synthetic scenes that match the scenes in a specific dataset \cite{Gaidon2016-zh,Kar2019-zn}. The objective is to reduce the burden of acquiring and labeling new training data.  This approach synthesizes images as a form of domain adaptation.

A third purpose is to co-design cameras and networks \cite{Blasinski2018-se,Liu2019-lc,Liu2019-ji}.  The goal of this application is to explore camera design that spans a wide range parameters, exceeding the consumer photography cameras, in order to discover systems whose images improve the accuracy of object detection networks. Soft prototyping and synthetic data is essential for co-design. This approach synthesizes images as a form of image system optimization.

\begin{table}[]
\caption{A summary of the papers using data synthesis methods to improve generalization for automotive scenes.
}
\label{table1}
\resizebox{\linewidth}{!}{
\begin{tabular}{|l|l|l|}
\hline
Method (authors)          & Data generation method                                                                                                       & Purpose                                                                                               \\ \hline
Tremblay 2018 & Synthetic (UE4)                                                                                            & Data augmentation                                                                                     \\ \hline
Ros 2016      & Synthetic (Unity)                                                                                                 & Data augmentation                                                                                     \\ \hline
Wrenninge2018 & Ray tracing                                                                                                             & Data augmentation                                                                                     \\ \hline
Alhajia 2017  & \begin{tabular}[c]{@{}l@{}}Adding virtual 3d model \\ into real images\end{tabular}              & Domain adaptation                                                                                     \\ \hline
Carlson 2019  & Synthetic (GTA)                                                                                            & Domain adaptation                                                                                     \\ \hline
Gaidon 2016   & Synthetic (Unity)                                                                                                 & Domain adaptation                                                                                     \\ \hline
Kar 2019      & Synthetic (UE4)                                                                                                   & Domain adaptation                                                                                     \\ \hline
Ours (ISETAuto)          & \begin{tabular}[c]{@{}l@{}}Physically-based \\ multispectral (ray-tracing) \\ simulations of camera images\end{tabular} & \begin{tabular}[c]{@{}l@{}}Data augmentation;\\ Co-design cameras \\ and neural networks\end{tabular} \\ \hline
\end{tabular}}
\end{table}

\section{Methods}
\subsection{Camera datasets}
We used public camera data sets from KITTI, CityScape, and Berkeley Deep Drive (BDD).  These are RGB image data shared in the form of PNG (KITTI, CityScape) or JPEG (BDD) files along with corresponding object labels.
\subsubsection{KITTI}
We analyzed images from the KITTI dataset\footnote{http://www.cvlibs.net/datasets/kitti/setup.php} \cite{Geiger2012-qu}, which includes images from two monochrome and two color cameras.  We selected the color images for object detection.  The KITTI dataset camera had a pixel size of 4.65 um and a lens with a 4mm - 8mm focal length based on a varifocal video lens \cite{Geiger2012-qu}. 
\subsubsection{BDD}
The BDD dataset\footnote{https://bdd-data.berkeley.edu/} \cite{Yu2018-wr} contains 80,000 labeled images and video clips.  It contains an additional 20,000 unlabeled images and videos for testing. The annotations include image level tagging, object bounding boxes, drivable areas, lane markings, and full-frame instance segmentation. The BDD data set includes JPEG images acquired using a variety of cameras and image processing chains.
\subsubsection{CITYSCAPE}
The CityScape dataset\footnote{https://www.cityscapes-dataset.com/} \cite{Cordts2016-sz} contains 5000 (3700 available) pixel level labeled images collected in street scenes from 50 different cities under daytime. CityScape does not provide camera information for their dataset. 

\vspace{5mm}
We selected 3700 images from each dataset, extracted bounding box labels from segmentation labels. The KITTI and CityScape data were collected during daytime on city streets.  Unless other comparisons were of interest, we matched the BDD data by selecting 3700 images with the metadata description of { “city street” and “daytime”}. 

\subsection{Synthetic datasets}
\subsubsection{SYNTHIA}
SYNTHIA \cite{Ros2016-ub} is a labeled image dataset for driving scenarios created using a game engine \cite{Unity_Technologies_undated-at}. The images (960 x 720) represent a virtual city and come with precise pixel-level semantic annotations for many different classes: car, building, etc.
\subsubsection{Synscape}
The Synscape dataset \cite{Wrenninge2018-ri} was created using ray-tracing. There are 25,000 images with two different resolutions.  We used the images with 1440x720. Each image is annotated with class, segmentation instance, and depth information.
\subsubsection{ISETAuto: Simulated scenes}
Simulated scene radiance data and sensor irradiance were generated for a collection of 4000 city scenes, using the ISET3d software\footnote{ https://github.com/iset/iset3d } \cite{Liu2019-lc}.  This software defines the positions of vehicles and pedestrians that match traffic statistics using the Simulation of Urban MObility (SUMO) software package\footnote{http://sumo.sourceforge.net/} \cite{Krajzewicz2012-mm}. The specific objects (vehicles, pedestrians, bicyclists) scenes and their surface properties were assembled by random sampling from a database of nearly 100 car, 3 bus, 80 pedestrians, 10 bicycles. The buildings, trees and other city features were positioned by an ISET3d algorithm that samples from city and suburban building collections including more than 200 buildings. 

The scenes are rendered using a modified version of the ray-tracing software described in Physically Based Ray Tracing (PBRT) \cite{Pharr2016-fl} and implemented as a Docker container\footnote{https://hub.docker.com/u/vistalab}. Containerization supports reproducibility by incorporating all the dependencies, and it also supports cloud-scale computing because many scenes can be produced at the same time using Kubernetes. The camera optical system is simulated with PBRT: the camera lens for the simulations we used in this paper was a wide angle (56 deg) multi-element design with 6 mm focal length. The on-axis point spread of the lens has a full-width at half maximum of approximately 1.5 um, making it suitable for simulation with a range of pixel sizes.
\subsubsection{ISETCam: Simulated cameras}
The ISETCam\footnote{https://github.com/iset/isetcam} software converts the spectral irradiance at the sensor into sensor digital values \cite{Farrell2012-nl}.  The software can simulate a range sensor parameters including varying pixel size, color filters, and image processing. The sensor electrical properties we simulated were based on the specifications of the MT9V024 sensor manufactured by ON Semiconductor, a sensor that is designed for automotive machine vision applications with options for monochrome, RGB Bayer and RCCC color filter arrays. The MT9V024 sensor has relatively high light sensitivity and signal-to-noise with a linear dynamic range of 55 dB.  We fixed the sensor dye size to be 3.6 x 8.6 mm. Simulating changes in pixel size varies the number of pixels inversely to the square of pixel size.

The exposure control was simulated using either (a) a center-weighted algorithm, concentrated on the image region in front of the car, (b) an exposure-bracketing using three different captures (2 ms, 4 ms, 8 ms), or (c) a global-weighted algorithm, using the entire image. For the single exposure case, the exposure duration was set so that the brightest region part of this center region produced a sensor voltage that was 90\% of the voltage swing but constrained to be no longer than 16 ms (60 frames per second). 

The physical accuracy of the ISETCam camera simulation has been evaluated in prior publications. Several independent studies report a close correspondence between measured and simulated sensor performance \cite{Chen2009-bj,Farrell2008-sc,Farrell2012-nl}. Additional validation is based on the use of physical principles: the optics is modeled in PBRT using physical principles of ray tracing and Snell’s Law \cite{Kolb1995-sh} and diffraction \cite{Freniere1999-ly}. To confirm the implementation we compared the optics simulations with these physical principles and the widely used commercial program, Zemax \cite{Lian2019-cj}. The surface rendering (PBRT) is based on ray tracing algorithms, with an accurate simulation of physics of light and its interaction with many types of surfaces including metals, diffuse reflectors, retro-reflective materials, and glass. The interaction function is known as bidirectional reflectance distribution function (BRDF), is derived based on the real-world experiments, and the function itself is validated by comparing the measured results \cite{Ward1992-in,Marschner1999-ra}. The three-dimensional models of the cars are derived from a 3D scan of a real car and the physical dimensions are accurate. The ambient light spectral characteristics were measured using a multispectral lighting capture system and are validated \cite{LeGendre2016-cj}.

Further details about how the ISETCam sensor models are created can be found in \cite{Liu2019-ji}. 
 
\subsection{Object detection network}
We use Mask R-CNN \cite{Facebook_undated-wz,He2017-bl} to detect and localize objects.  MaskRCNN includes a region proposal network (RPN) that creates different regions (rectangles) on an image where an object might be found and another network to detect objects using these regions. In this paper, we evaluate datasets using only the bounding box output of Mask R-CNN. We choose ResNet 50 with FPN as the backbone of the network. We chose the anchor sizes for the RPN to be [32, 64, 126, 256, 512] with three different ratios: [0.5, 1, 2]. We trained and evaluated the model for cars and pedestrians using 4 Nvidia P100 GPUs. We report the results for cars but not pedestrians. The car models are highly realistic while the human models are less compelling.  Accurate human models are needed for a precise evaluation of the pedestrian class. 

\subsection{Metrics}
\subsubsection{Object detection system performance}
When the network identifies an object within a bounding box, and that box overlaps with at least half of the area of the bounding box on the labeled pixels, the detection is considered correct. Combining the hits and false alarms from this measure, we obtain the average precision based on the intersection over union, a metric that is widely used in machine-learning \cite{Everingham2019-qm}. The AP is equivalent to measuring the area under the receiver operating characteristic defined in signal detection theory \cite{Swets1978-vb}. Unless indicated otherwise, we use the shorthand AP (Average Precision) to describe AP@0.5IOU.
\subsubsection{Labeling policy}
Different camera datasets use different labeling policies.  Although the datasets are annotated by humans, KITTI annotates the 2d bounding box with a minimum requirement of height of the box to be larger than 25 pixels, whereas BDD and CityScape contain objects with smaller bounding boxes labeled.

We use two different labeling policies for our evaluation. The experiments conducted on pixel size generalization contains the labels of all the cars within 150 meters regardless of the pixel size.  For the other evaluation experiments, we labeled the cars using the KITTI labeling policy with a minimum box size.
\Figure[t!](topskip=0pt, botskip=0pt, midskip=0pt)[width=3 in]{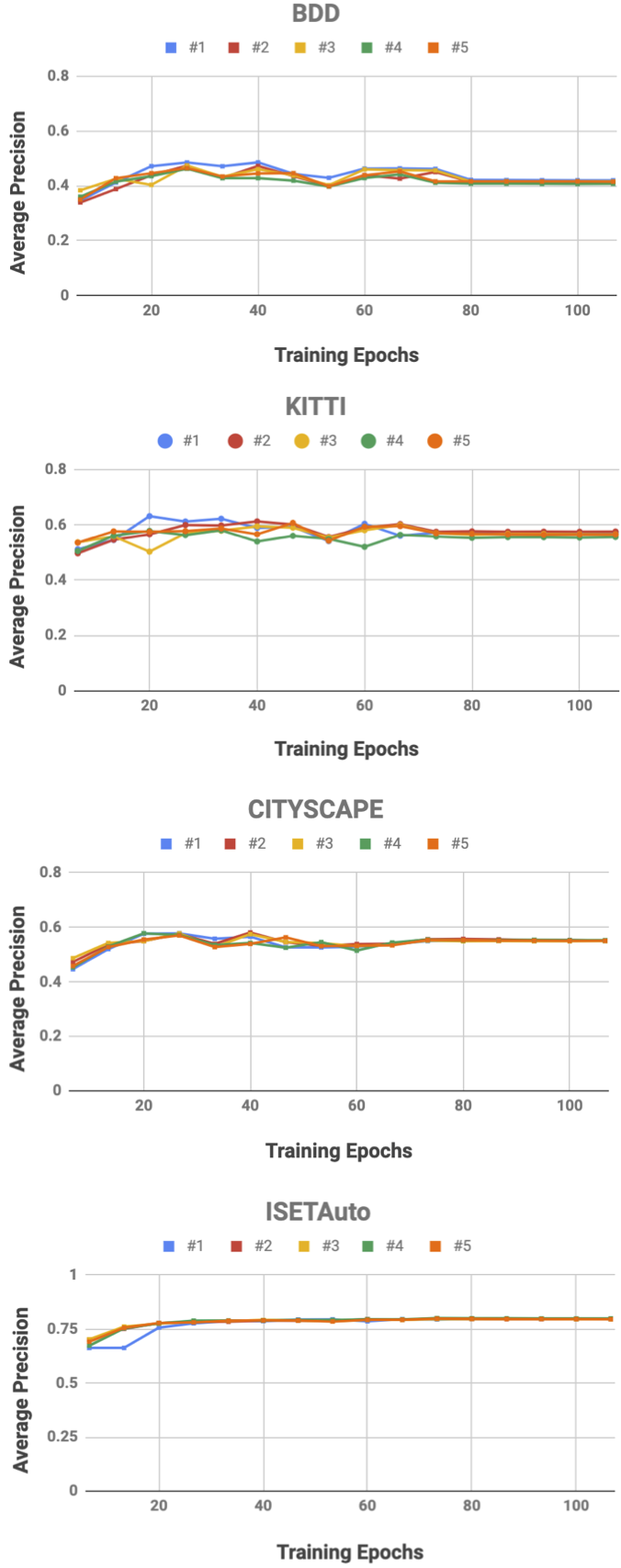}
{Replicability of model training at multiple checkpoints.  A ResNet model was trained on ISETAuto data.  The four panels show how AP performance changed when testing on four different datasets (the four panels), each run three times (three curves).   In all cases the stable values were reached after 80 epochs, and the standard deviation of the AP after stabilization is below 0.01.
\label{fig1}}
\Figure[t!](topskip=0pt, botskip=0pt, midskip=0pt){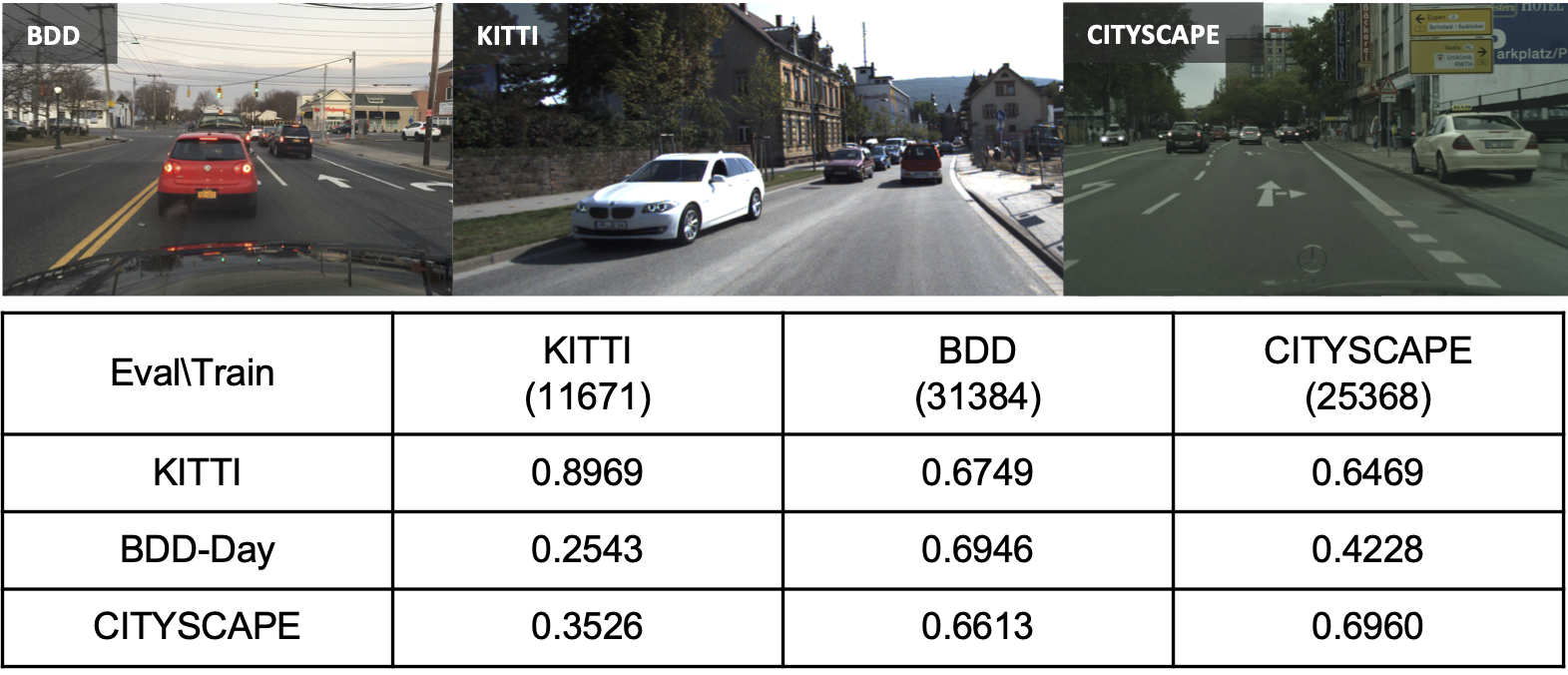}
{\textbf{Top}. Representative scenes from three different public datasets used for autonomous driving.  Images from different datasets are easily distinguished by human subjects. The discrimination can be made in large part by color cast and contrast.  \textbf{Bottom}. The table shows the AP@0.5IOU measured when training on 3000 images from the data set listed in the column header, and evaluating with the images from the data set listed at the left of the row. The column headers show the number of cars in  each of the data sets. The values in parentheses in each column are the number of car objects in each of the data sets.
\label{fig2}}
\subsubsection{Object detection network training}
We trained all models from scratch. Training was based on 3000 images which were presented to the network with a batch size of 8 images per training step; model weights were updated after each batch.  For example, for the case of 40,000 training steps, a total of 320,000 images were presented so that the training set of 3000 images was presented about 106 times (epochs). The model was evaluated and the AP values saved at 16 checkpoints. Model performance was evaluated based on 700 images that were not used in training (held out). 

We evaluated how closely repeating the training returns the same AP value (Figure 1). We initialized the ResNet50 network randomly and measured AP through 106 training epochs for four different cases: training on the ISETAuto synthetic data and evaluating on the empirical (KITTI/BDD/CITYSCAPE) and ISETAuto images. The five curves track the change in AP performance for different replications. For the empirical data, there is variation in the AP level between replications up to about 80 training epochs. As training continues, the AP level settles near the peak value with very little variation;  the standard deviation of the AP value is well below 0.01 in all conditions.  Using the synthetic ISETAuto data for training and testing, the AP level stabilizes after approximately 80 training epochs.  Hence, the assessments in this paper report AP values after 80 training epochs, after the AP value stabilizes. Based on the analysis in Figure 1, we consider AP values to be reliable within +/-0.01.

\section{Experiments}
We begin by analyzing generalization between camera datasets. We use the generalization between camera datasets as a baseline to compare generalization between synthetic and camera datasets. Finally, using the soft prototyping tools we examine generalization between images obtained using different cameras and post-processing methods.
\subsection{Generalization between camera datasets}
\subsubsection{Global generalization}
\Figure[t!](topskip=0pt, botskip=0pt, midskip=0pt){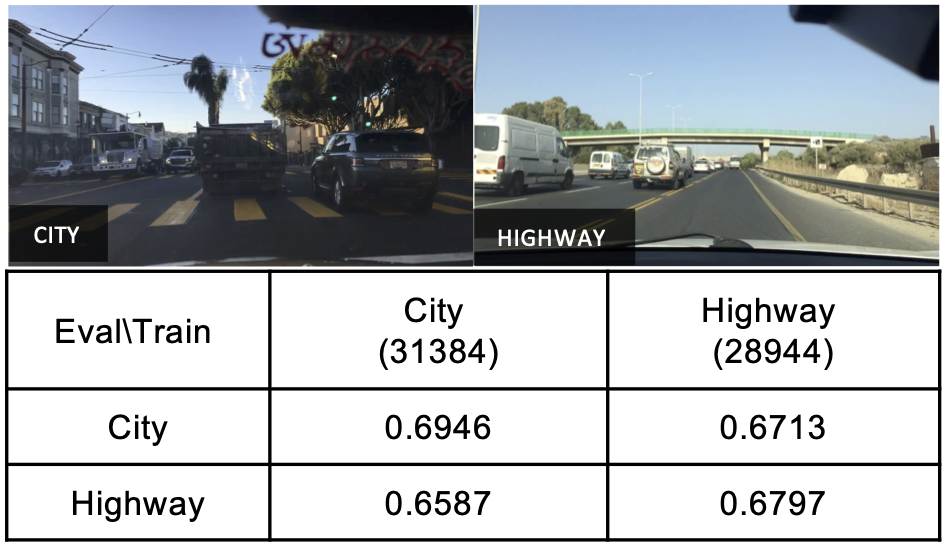}
{\ Generalization between City and Highway scenes in the BDD dataset. Other details as in Figure 2.
\label{fig3}}
\Figure[t!](topskip=0pt, botskip=0pt, midskip=0pt){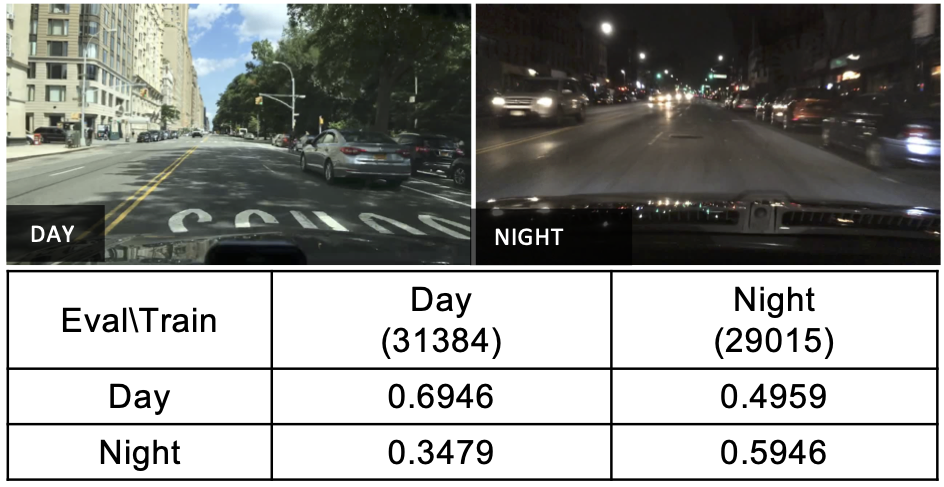}
{Generalization between different times of day for images in the BDD dataset. Other details as in Figure 2.
\label{fig4}}
\Figure[t!](topskip=0pt, botskip=0pt, midskip=0pt){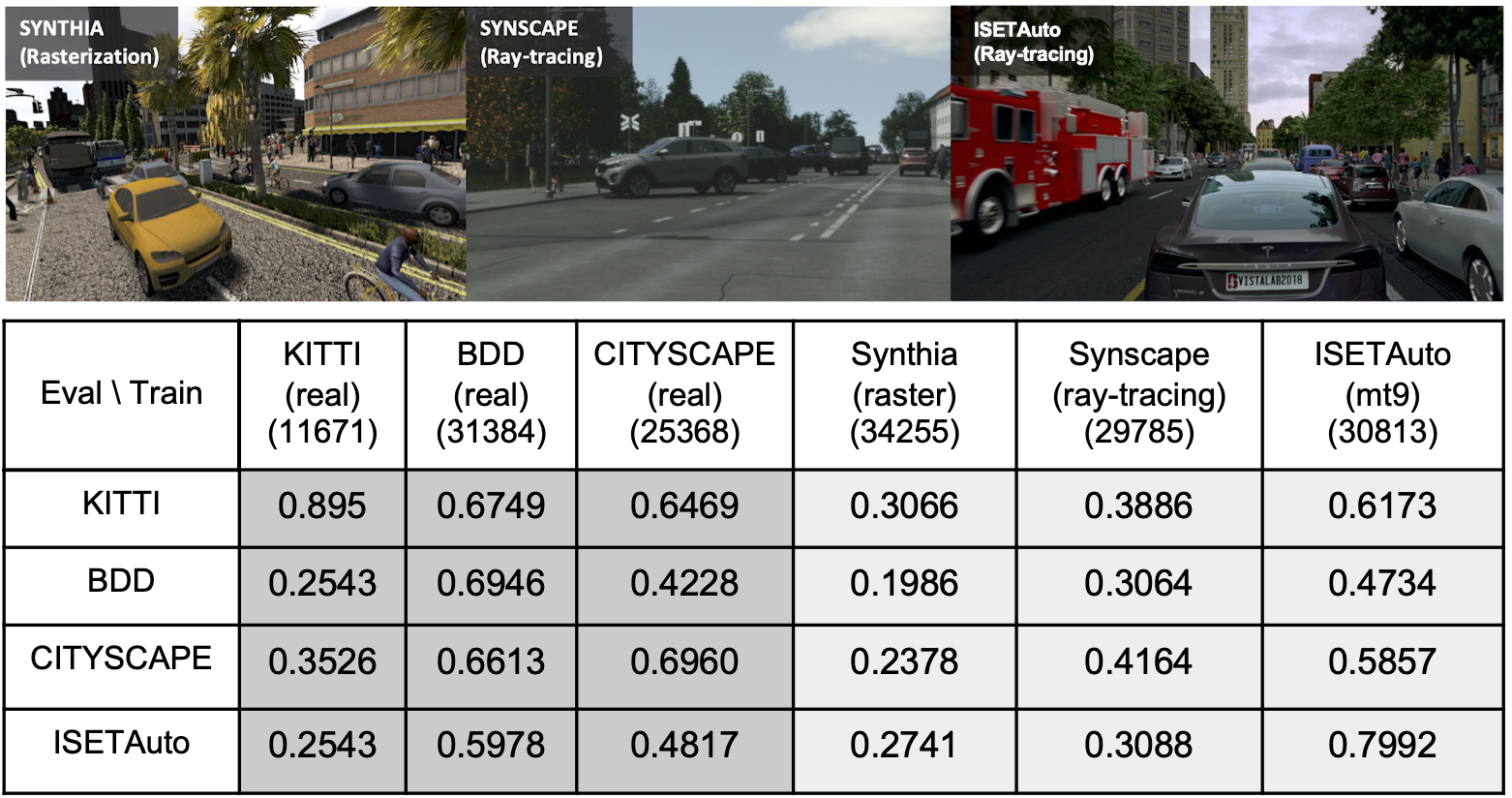}
{\textbf{Top}. Representative scenes from three different synthetic datasets used for autonomous driving.  SYNTHIA used a rasterization method to generate the scenes, Synscpae and ISETAuto used ray-tracing. The quality of the images from SYNTHIA is less physically accurate than the images from Synscape and ISETAuto, but they can be rendered faster.  \textbf{Bottom}. The table shows the AP@0.5IOU measured when training on 3000 images from the data set listed in the column header, and evaluating with the data set listed at the left of the row. The column headers show the number of cars in  each of the data sets.
\label{fig5}}
We trained the network three times, using 3000 labeled images from KITTI, BDD or CITYSCAPE.  After training, we evaluated (350 images for validation at each checkpoint, and another 350 images to measure system performance) and calculated the AP for car detection using held-out data from KITTI, BDD, CITYSCAPE. The images in Figure 2 illustrate the appearance of typical images from three empirical data sets. 

The numerical values in Figure 2 shows the AP for each of the training and evaluation conditions. The first column shows that training on the KITTI dataset achieves a high performance level, but generalizes poorly to the BDD and CITYSCAPE data sets.  The second column shows that BDD training generalizes well to the other datasets.  The third column shows that CITYSCAPE generalizes to KITTI well, but does not generalize well to BDD. A few reports of generalization scores in the literature are consistent with these values.  In one study, training on 70,000 BDD images achieved a generalization to KITTI of about 0.55 using a criterion of AP@0.7IOU  \cite{Prakash2018-ep}.  With 3,000 BDD images, we found a generalization performance level of 0.67 at AP@0.5IOU.

A particularly clear conclusion from examining the generalization table is this:  The KITTI data are an outlier.  Training on the KITTI dataset does not generalize to the BDD or CITYSCAPE data, and training on the BDD or CITYSCAPE does not generalize well to the KITTI data. The difference in generalization may be due in part to the use of different cameras, and it may also be explained in part by labeling policy differences, which results in different number of instances for model training. The KITTI data only labels cars when the bounding box is at least 25 pixels on a side, while the other datasets include cars with smaller bounding boxes.  Given that there are no small bounding boxes labeled in the KITTI data set, generalization performance levels on the BDD and CITYSCAPE data will be low. 

There is a significant asymmetry in the generalization values. Training on BDD generalizes well to CITYSCAPE, but training on CITYSCAPE does not generalize well to BDD.  The difference may be due to the diversity of camera types used to collect the data for the BDD sample, while only a single type was used for CITYSCAPE. 

\subsubsection{Scene type generalization}
The type of scene is another limit on generalization.  The BDD images include labels identifying images as arising from cities, highway, day, night, clear and rain.  We examined generalization between training on city and highway scenes (Figure 3).  Trained on highway generalizes well to city (less than 1\% variation); trained on city has a  4\% performance drop to highway scenes.  This is surprising because normally we assume the model can learn more features when it is trained on a complex background (city scenes are more complex than highway); yet the generalization is better in the opposite direction. 

Training on images labeled as Day generalizes poorly to Night, and similarly training on Night generalizes poorly to Day (Figure 4).  This is unsurprising because the statistics of these images are so different from one another. We also noticed that the Night images are more difficult for the model to learn than Day images (lower performance with the same training setting. The features learned in Night images generalize better to Day images than the reverse. These values set a baseline for expectations between conditions with poor generalization. 

\subsection{Generalization between synthetic and camera data sets}
\subsubsection{Global generalization}

Next, we trained and evaluated using three types of synthetic images.  The SYNTHIA images \cite{Ros2016-ub} are generated using a rasterization method; Synscape \cite{Wrenninge2018-ri} and ISETAuto \cite{Liu2019-lc,Liu2019-ji} are generated using ray tracing. We compared the generalization between the synthetic and camera data as well as the generalization between the synthetic data sets (Figure 5).

Training using the ISETAuto images generalizes better than training with either of the other synthetic data sets. Training on the ISETAuto data generalizes to the KITTI dataset (0.62) only slightly less than training on the BDD (0.67) and CITYSCAPE (0.65) image data. Training on the ISETAuto generalizes to BDD (0.47) slightly better than training on CITYSCAPE (0.42) or KITTI (0.25). Training on ISETAuto generalizes better to CITYSCAPE  (0.59) than training on KITTI (0.35).  Generalization from SYNTHIA and Synscape to any of the other data sets is poor, including the three public empirical data sets and ISETAuto.

\subsubsection{Object distance generalization}
We examined whether failures to generalize are restricted to the smaller and more distant cars by calculating AP@0.5IOU as a function of distance. Specifically, we evaluated the performance of networks trained on different public and synthetic datasets on the synthetic images in the ISETAuto, which include metadata describing the distance to each of the cars.  Failures to generalize were present for cars at all distances (Figure 6).  This suggests that the limited generalization is not associated with the distance or size of the car in the image.
\Figure[t!](topskip=0pt, botskip=0pt, midskip=0pt)[width=3 in]{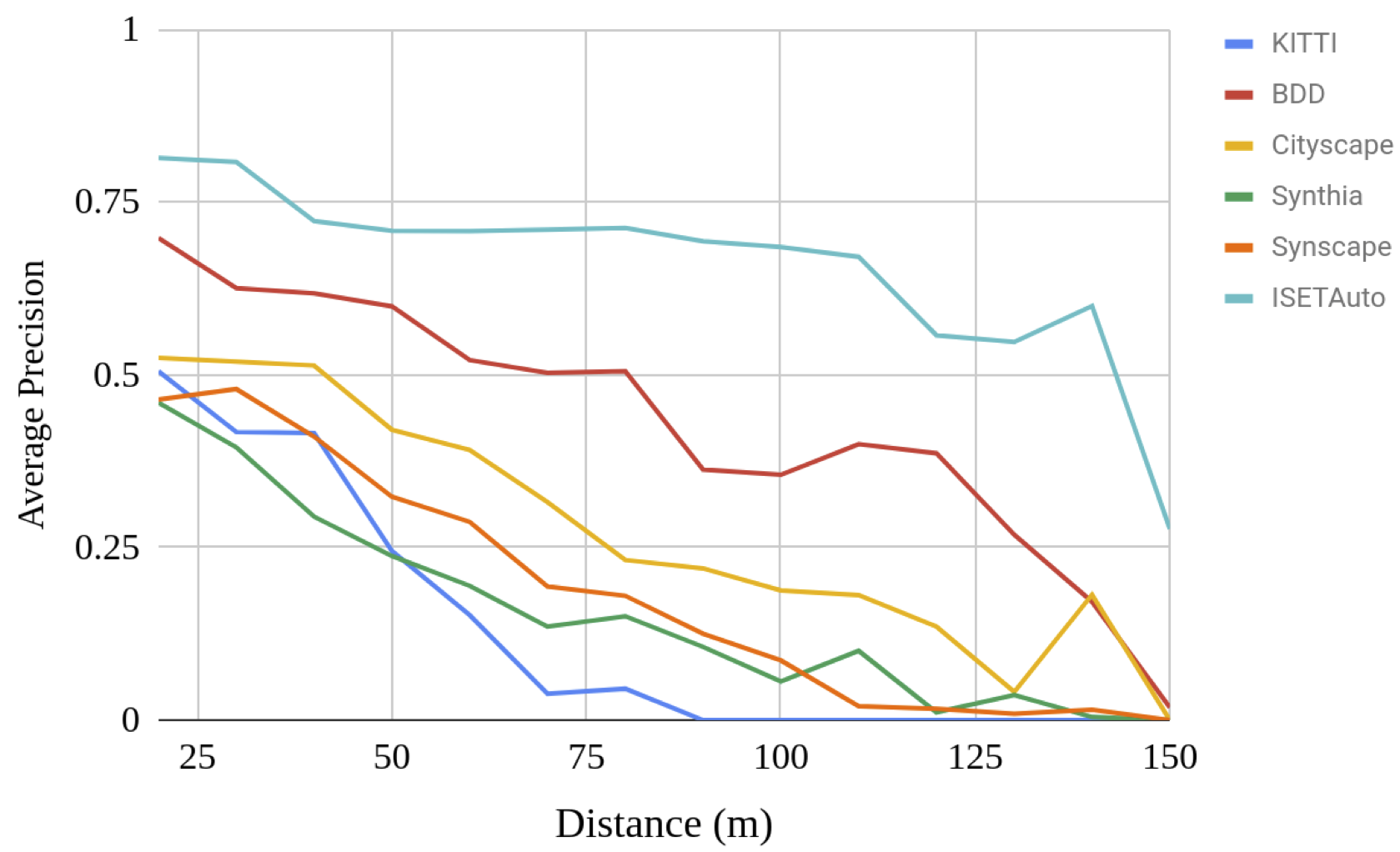}
{Performance (AP@0.5IOU) for a network model trained on different real and synthetic datasets (legend).  The networks are evaluated using the ISETAuto dataset which includes metadata about the distance to each car \cite{Liu2019-ji}. The ordering of the generalization remains consistent for cars at all distances. 
\label{fig6}}
\subsubsection{Sensor pixel size generalization}
The three public datasets contain images with different camera spatial resolutions.  To examine the impact of pixel size on generalization, we synthesized ISETAuto images with sensors at pixel sizes ranging from 1.5 um to 6 um; each training set had the same number of objects (Table 2). The ISETAuto images synthesized at 1.5 um generalize the best to CITYSCAPE, which has the largest number of image spatial samples.  The ISETAuto data set at 3 um generalizes the best to BDD.  The ISETAuto images synthesized using 3 um to 6 um pixel sizes generalize at similar levels to the KITTI data. 
\begin{table}[]
\caption{Generalization is assessed when training with an ISETAuto dataset at a range of pixel sizes (columns).  Evaluation was performed with the public datasets (rows). The shaded box in each row shows the pixel size that generalized best to each of the public datasets (maximum within the row). The KITTI dataset was acquired using a camera with a 4.65 um pixel.  The BDD and CITYSCAPE data cameras are not specified.
}
\centering
\begin{tabular}{|c|c|c|c|c|}
\hline
Eval \textbackslash Train                                                           & \begin{tabular}[c]{@{}c@{}}1.5 um\\ (2546x1188)\\ (31384 )\end{tabular} & \begin{tabular}[c]{@{}c@{}}3 um\\ (1268x594)\\ (31384 )\end{tabular} & \begin{tabular}[c]{@{}c@{}}4.5 um\\ (950x446)\\ (31384 )\end{tabular} & \begin{tabular}[c]{@{}c@{}}6 um\\ (634x298)\\ (31384 )\end{tabular} \\  \hline
\begin{tabular}[c]{@{}c@{}}\\[0.01cm]KITTI\\ (1224x370)\\ (11671)\end{tabular}      & 0.5499                                                                           & 0.5740                                                                        & \cellcolor{gray}0.5829                                                        & 0.5715                                                                       \\ \hline
\begin{tabular}[c]{@{}c@{}}\\[0.01cm]BDD\\ (1280x720)\\ (31384 )\end{tabular}        & 0.4146                                                                           & \cellcolor{gray}0.4443                                                                        & 0.4085                                                                         & 0.3602                                                                       \\ \hline
\begin{tabular}[c]{@{}c@{}}\\[0.01cm]CITYSCAPE\\ (1920x1080)\\ (25368 )\end{tabular} & \cellcolor{gray}0.5750                                                                           & 0.5465                                                                        & 0.5301                                                                         & 0.4498                                                                       \\ \hline
\end{tabular}
\end{table}
\subsection{Generalization between synthetic data sets}
\subsubsection{Sensor pixel size generalization}
We compared the generalization between networks trained with different pixel sizes (Figure 7). Performance generally declines as pixel size increases, as expected: sensors with larger pixels have lower spatial resolution.  The generalization measurements suggest that training a network with pixels of size S generalizes well to smaller pixel sizes, say S/2. The training generalizes poorly, however, to larger pixel sizes.
\Figure[t!](topskip=0pt, botskip=0pt, midskip=0pt){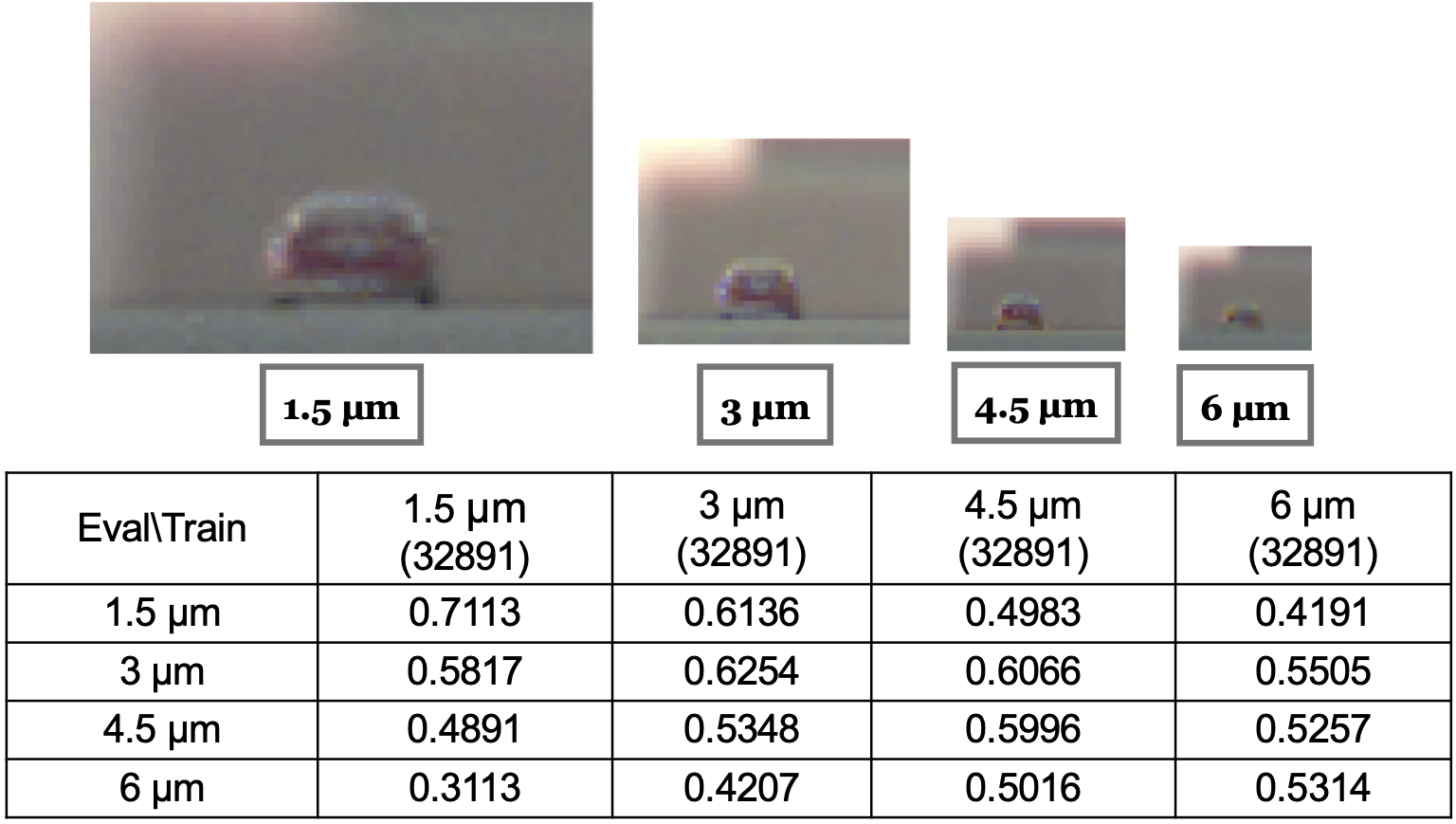}
{Generalization is assessed between networks trained and evaluated with datasets with different pixel sizes (ISETAuto).  The sensor size (field of view) is constant as we vary pixel size, so that larger pixels have fewer spatial samples (images at the top). Upper image shown a car at 150 meter from a pinhole camera (FOV 45 degree) simulated with different pixel sizes. Other details as in Figure 2.
\label{fig7}}
We find it surprising that training on a 6 um pixel generalizes very well to a 3 um pixel.  In fact, the performance is slightly higher.  This may reveal a tradeoff between the precision of the training and the quality of the image data.  The 6 um sensor often has very small bounding boxes (measured in pixels) for the cars. Thus the limited generalization may be counter-balanced by the improvement in the image resolution.  
\subsubsection{Color filter array generalization}
For car detection the choice of color filters - ranging from monochrome to RGB to RCCC - generalizes well (Figure 8).  A network trained on a monochrome sensor generalizes very well to either the raw sensor data from an RGB or RCCC sensor. Apparently the network learns to extract information that is not eliminated by the variations introduced by the color filters. Moreover, the generalization is symmetric in the sense that training on any of the different sensors evaluates well on data from the other sensors.

It may be that detection of other objects (e.g., traffic lights) may benefit from choosing the color wisely.  The simulations for cars, however, suggests that a monochrome sensor is no worse than a color sensor.
\Figure[t!](topskip=0pt, botskip=0pt, midskip=0pt){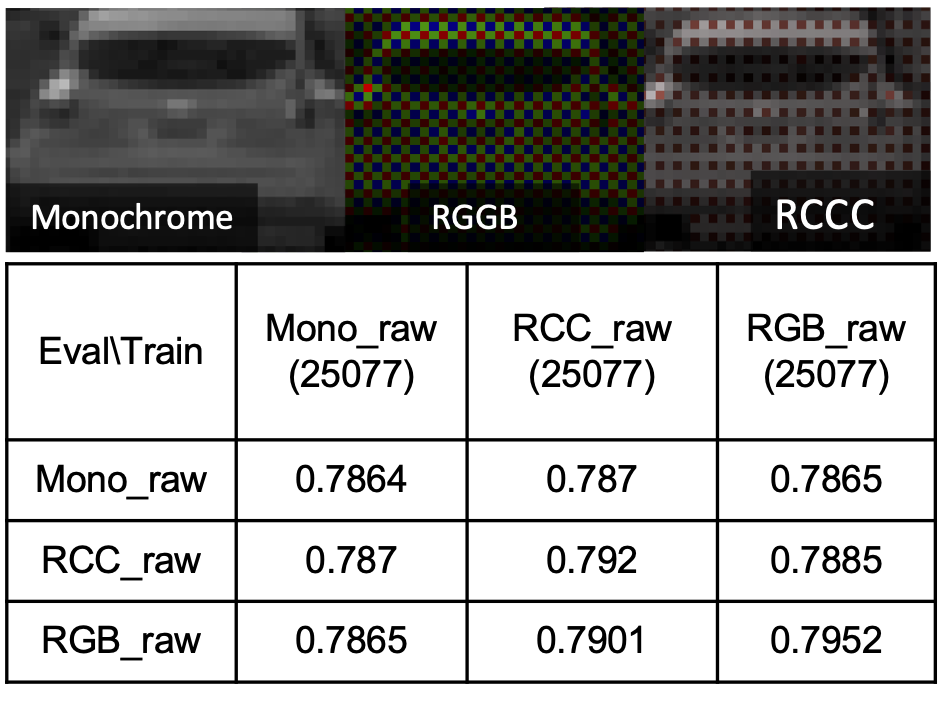}
{Generalization is assessed between networks trained and evaluated using 10 bit sensor raw images with different color filters (ISETAuto simulations). When using 10 bit pixels, generalization is very high. Other details as in Figure 2.
\label{fig8}}
\begin{table}[]
\caption{Generalization is assessed on monochrome sensor raw images with different bit-depth}
\centering
\resizebox{\linewidth}{!}{
\def\arraystretch{1.5}
\begin{tabular}{|c|c|c|}
\hline
Eval \textbackslash Train & \begin{tabular}[c]{@{}c@{}}Mono\_raw\_8bit\\ (25077)\end{tabular} & \begin{tabular}[c]{@{}c@{}}Mono\_raw\_10bit\\ (25077)\end{tabular} \\ \hline
Mono\_raw\_8bit & 0.7864 & 0.7171 \\ \hline
Mono\_raw\_10bit & 0.623 & 0.7864 \\ \hline
\end{tabular}}
\end{table}
\Figure[t!](topskip=0pt, botskip=0pt, midskip=0pt){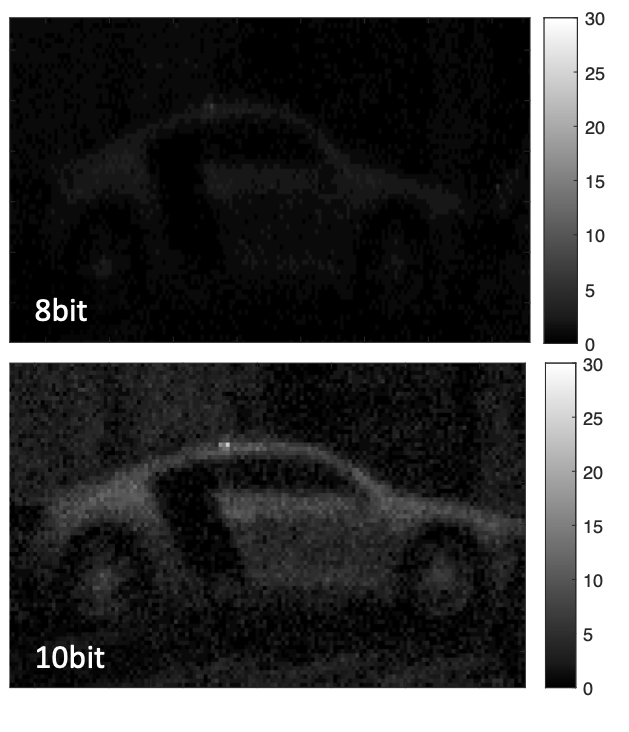}
{A car captured in the dark part of a scene with two different cameras. The scene has a bright sky that effectively sets the exposure duration. The cameras with the two different bit-depths do not represent the car equally well. The 8 bit camera (top) lacks detail, while the 10-bit camera (bottom) has enough intensity resolution to contrast the car from its surroundings.
\label{fig9}}
\subsubsection{Bit-depth generalization}
Figure 9 shows a car inside a shadow of a bright scene. The critical information about the car is represented in the smaller digital values (0-30).  For the 8-bit camera (top) the lowest 30 levels have a very poor representation of the car, which is difficult to see. In these same levels, the 10-bit camera (bottom) includes a reasonable representation of the same car. Note that changing the bit-depth does not change the pixel dynamic range; it increases the contrast within the dark digital levels. The simulations were performed using a center-weighted exposure algorithm.

For a monochrome sensor there is little performance difference between networks trained and evaluated on 8-bit (0.79) and 10-bit (0.79) (Table 3).  However, the networks trained at different bit-depths do not generalize well. Training on 8-bit  evaluates on 10-bit images at a low level (0.62), and training on 10-bit evaluates on 8-bit at a low level (0.71). 

For RGB sensors the generalization across bit-depths is even worse (Table 4).  Training with an 8-bit RGB sensor reaches a low performance level (0.38); increasing to a 10-bit quantization produces a much higher AP@0.5IOU (0.79).  Naturally, the generalization between 8- and 10-bit is very poor. 

For the RGB sensors, automotive scenes, and a center-weighted exposure algorithm, we find that increasing to a 16-bit depth is not significantly different from 10-bit.  In each case, however, the generalization is modest or poor.  Training on 10-bit evaluates at 0.71 on 16 bit, and training 16 bit evaluates at 0.68 on 10 bit.  Hence, the network is using the number of levels as an important signal.

\begin{table}[]
\caption{Generalization is assessed on RGB sensor raw images with different bit-depth.
}
\centering
\resizebox{\linewidth}{!}{
\def\arraystretch{1.5}
\begin{tabular}{|c|c|c|c|}
\hline
Eval \textbackslash Train & \begin{tabular}[c]{@{}c@{}}RGBraw\_8bit\\ (25077)\end{tabular} & \begin{tabular}[c]{@{}c@{}}RGBraw\_10bit\\ (25077)\end{tabular} & \begin{tabular}[c]{@{}c@{}}RGBraw\_16bit\\ (25077)\end{tabular} \\ \hline
RGBraw\_8bit & 0.375 & 0.2128 & 0.091 \\ \hline
RGBraw\_10bit & 0.0965 & 0.7901 & 0.6938 \\ \hline
RGBraw\_16bit & 0.034 & 0.7184 & 0.7923 \\ \hline
\end{tabular}}
\end{table}
\Figure[t!](topskip=0pt, botskip=0pt, midskip=0pt){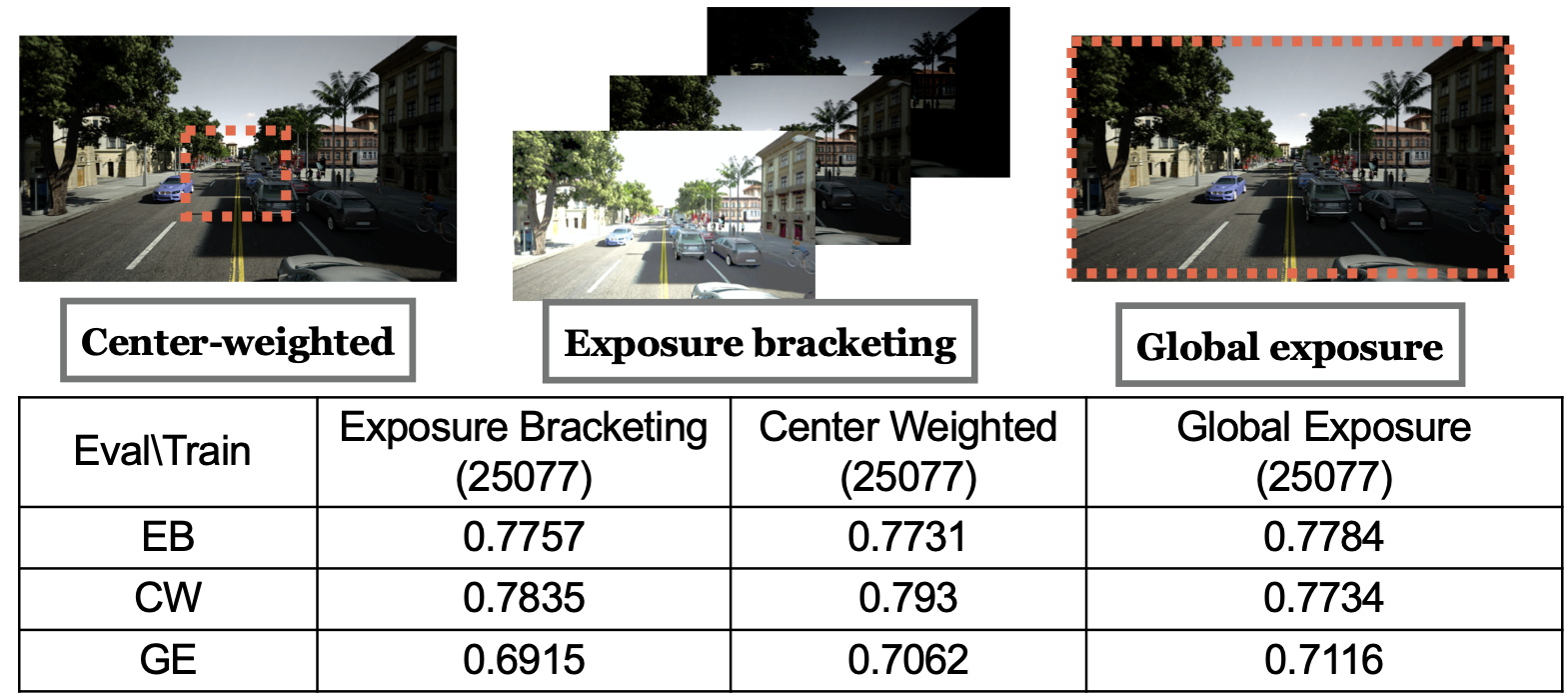}
{Generalization between different exposure-control algorithms (Exposure bracketing, EB;  Center-weighted, CW; Global exposure, GE). On the whole EB and CW are similar, though they can differ in important edge cases \cite{Liu2019-ji}.  The GE performance is low, presumably because it is a poor method for capturing data.  But training on EB generalizes well to EB and CW showing that what is learned is similar. Other details as in Figure 2.
\label{fig10}}
\subsubsection{Demosaicking interacts with bit-depth}
Demosaicking matters a great deal at 8 bits, but less so at 10 bits.  When training and testing with the raw sensor image at 8 bits, performance is very poor (0.375, Table 4).  Demosaicing the 8-bit sensor data significantly improves the performance (0.7997).   Surprisingly, demosaicking does not significantly improve performance for a sensor with 10 or 16 bits.
 
The value of demosaicing for 8-bit images has been previously reported \cite{Buckler2017-iy}.  Their methods did not permit them to check the dependence on bit depth. 

\subsubsection{Sensor control algorithms (Exposure)}
The exposure control algorithm is very important for high dynamic range scenes, such as the ones found in driving conditions.  If the exposure is set too short significant portions of the image may under-exposed and be very dark; if the exposure is too long significant portions of the scene may be saturated.

We compared three different exposure algorithms (Figure 10). The global exposure algorithm sets the exposure based on the brightest portion of the image, and this algorithm has the lowest performance.  The very bright sky, or large specular regions, choose a short duration and as a result much of the image may be too dark for car detection.

Restricting the exposure to a central part of the scene is typically a better sample of the important parts of the image for detection, and this improves performance. Exposure bracketing algorithms effectively increase the sensor dynamic range in exchange for the time required to obtain multiple captures. 

There is good generalization between networks trained on global exposure, center-weighted and exposure bracketing.  This suggests that training on any of these algorithms teaches the network the same critical features, and the performance differences arise because the global exposure data are poor quality.
 
\subsubsection{Post-acquisition processing (Gamma correction)}
Post-acquisition processing algorithms also influence performance and limit generalization (Figure 11).  We compared the performance of networks trained using the raw sensor values with a network trained using the sensor values passed through an exponential (gamma) function, and with a network trained in which the gamma value is set adaptively depending on the image contents.

The absolute network performance is fairly robust to the gamma value:  performance levels are roughly equal for a range of exponents (diagonal entries of Figure 11).  However, generalization is poor when trained with different gamma values.  The generalization between different gamma levels between very close values (0.2 and 0.3) is good.  But post-processing exponents that differ by more than 0.1, say (0.1 vs 0.3) or (1 and 0.3), generalize poorly.
\Figure[t!](topskip=0pt, botskip=0pt, midskip=0pt){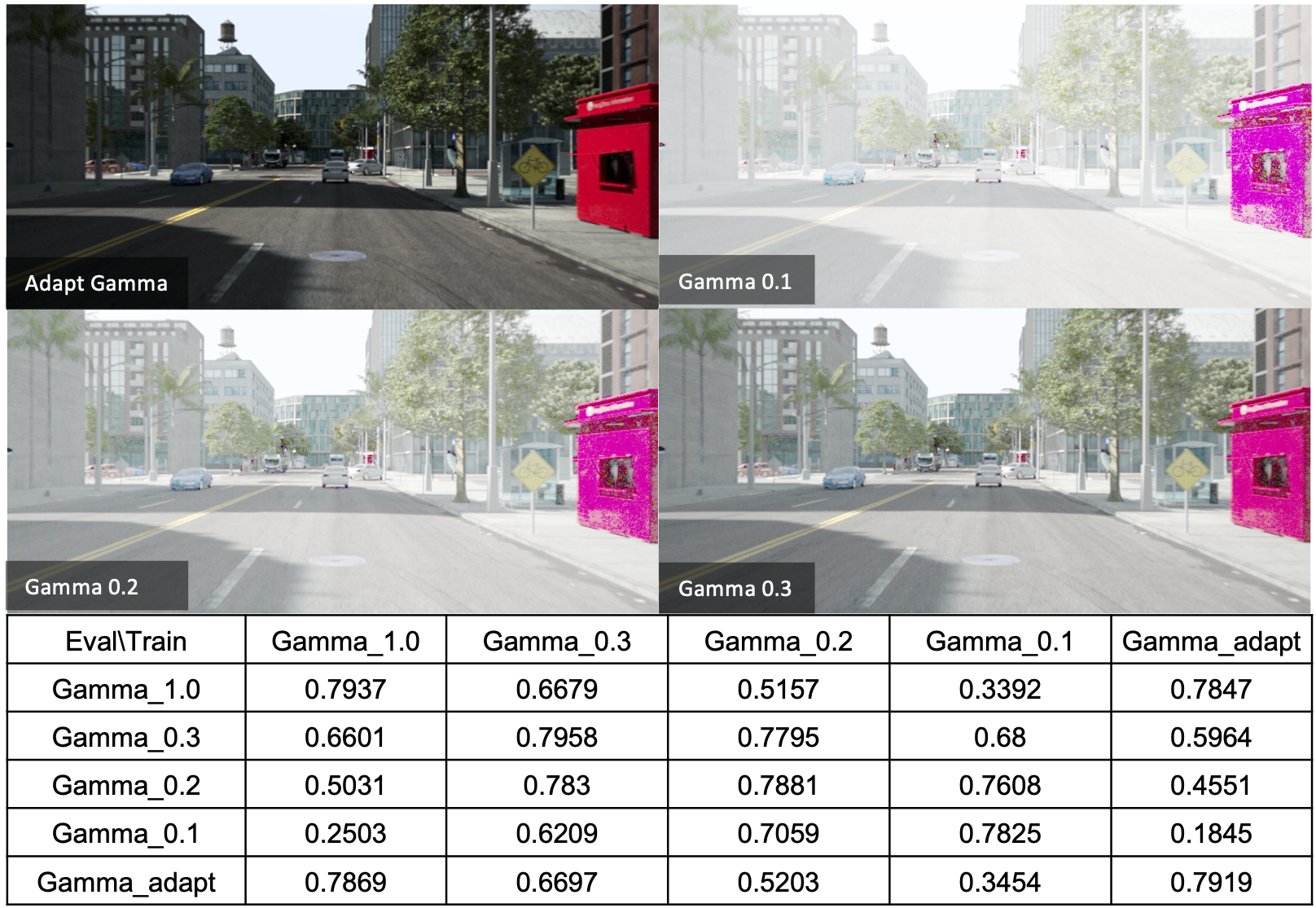}
{Generalization is assessed between networks trained and evaluated using sensor raw images with different color filters (ISETAuto simulations).  When using 10 bit pixels, generalization is very high. Other details as in Figure 2.
\label{fig11}}

\section{Discussion and summary}
There is widespread agreement that there are limitations to network generalization \cite{Torralba2011-pd}.  There is also consensus that modern artificial networks do not generalize as well or in the same way as humans \cite{Geirhos2018-yg}.  There are conflicting views as to whether generalization is specifically poor between simulations of camera images and camera images \cite{Carlson2019-oe,Movshovitz-Attias2016-bg,Sun2014-mx,Vazquez2014-vp,Tremblay2018-az,Ros2017-rs}. 

This paper quantifies generalization between different types of image data. The results clarify that performance derived from training on physically-based multispectral simulations of camera images generalizes to camera images at nearly the same level as performance generalization between different camera image datasets. 

\subsection{Factors limiting generalization}
Using simulation methods we analyzed the impact of camera parameters and post-acquisition processing on generalization. It is difficult to perform analyses at this level of granularity from public data sets of camera images. For example, the BDD dataset includes a large number of images of a wide range of scenes, obtained using cameras with different lenses, sensors, acquisition, and post-acquisition algorithms.  Separating out the impact of one or even a combination of camera parameters from this complex dataset is impossible.  Obtaining a new large and more controlled dataset of real camera images is impractical.  It is feasible, however, to create synthetic data that systematically vary individual camera parameters, and we have done so.  These experiments quantify generalization with respect to parameters including pixel size, exposure control, color filters, bit-depth, and post-acquisition processing. 

In related work we describe technical methods for creating and reasons for using synthetic data. Table 5 lists results from the literature that use these methods to train networks and assess performance on public data sets (KITTI and CITYSCAPE).  The physically-based multispectral simulations of camera images (ISETAuto) performs at about the same level as data augmentation methods, domain randomization, and domain adaptation (style transfer). Several of these methods generalize to the KITTI data about as well as training on BDD (0.67) or CITYSCAPE (0.65).  

The last three rows of the table show that the absolute level of performance with ISETAuto has not reached its highest possible level. In row 5 we trained with 3,000 images containing 25,077 objects and reached an AP level of 0.569 on Cityscape.  Adding more objects (29,942) increased performance to 0.586.  Adding 1,000 more images for a total of 36,885 objects further increased performance to 0.625.  We found the same pattern of increasing improvement using the KITTI test. This level of generalization is very similar to the generalization between camera images.  We summarize Table 5 by noting that the quality of the synthetic images has reached a level where they can be helpful for several different objectives, including camera design.

\begin{table*}[t]
\caption{Generalization using networks trained with simulated data and tested on camera images. Rows 1-4 are results taken from the literature.  Rows 5-7 are from simulations with ISETAuto. All performance levels were measured using AP@0.5IOU.}
\centering
\resizebox{\textwidth}{!}{
\def\arraystretch{1.5}
\begin{tabular}{|c|c|c|c|c|}
\hline
Training on & Number of training images & Detection network & Test on KITTI & Test on CITYSCAPE \\ \hline
Alhaija2017 & 4k & Faster RCNN using VGG pre-trained on ImageNet & 0.51 & 0.37 \\ \hline
Tremblay2018 & 10k & Faster RCNN trained from scratch & 0.58 & ------- \\ \hline
Kar2019 & 4k & MaskRCNN-Resnet-50-FPN pre-trained on ImageNet & 0.66 & ------- \\ \hline
Carlson201 & 10k & FasterRCNN (might be from scratch) & 0.52 & 0.35 \\ \hline
Ours (ISETAuto) & 3k (25077 instances) & MaskRCNN trained from scratch & 0.604 & 0.569 \\ \hline
Ours (ISETAuto) & 3k (29942 instances) & MaskRCNN trained from scratch & 0.617 & 0.586 \\ \hline
Ours (ISETAuto) & 4k (36885 instances) & MaskRCNN trained from scratch & 0.64 & 0.625 \\ \hline
\end{tabular}}
\end{table*}
\subsubsection{Generalization and similarity}
Data from many of the Tables show that generalization between data sets is asymmetric.  For example, training on KITTI generalizes poorly to BDD, but training on BDD generalizes well to KITTI.  The asymmetry in generalization is sensible:  one data set may span a much greater range and be more instructive.  

Some authors have suggested that to improve generalization one might create image data sets that are more similar. We can measure the distance between image collections in various ways, including the Kernel Inception Distance (KID) score \cite{Binkowski2018-gb}. For some applications there may be value in using generative adversarial networks to reduce the distance between image data sets. But note that by definition distance is a symmetric measure, and for this reason it is not a replacement for generalization which can be (and is) asymmetric.  

We computed the KID values between the image data sets used in our analyses.  The KID distances are roughly consistent with generalization; for example, the SYNTHIA data set is the least similar (most distant) to the others.  Synscape and  ISETAuto are similar to KITTI and Cityscape, but ISETAuto is more like BDD.  The greatest similarity is between ISETAuto and Cityscape. These distances are roughly consistent with the generalization, but of course they do not capture the asymmetry.

\subsection{future}
The generalization between the ray-traced synthetic images and the camera images is encouraging. The performance level using synthetic images for the car detection application is already adequate to be helpful in advancing all three objectives (universal network design, specific camera simulation, co-design of camera and network). It is likely that more experiments and innovation can increase the level of generalization.  

For our project, camera and network co-design, many parameters remain to be explored. These include camera placement, lens designs, and novel sensor designs.  The quantification of generalization we report here suggests that a soft prototyping tool will provide useful guidance to explore this parameter space.

The simulation environment also enables us to control parameters of the scene, in particular object placement. This provides an opportunity to perform additional experiments in which objects are added to the scene in unusual locations (e.g., \cite{Tremblay2018-az}).  The necessity of accurately placing objects for training is an interesting and open question.

Finally, one might aspire to extend this work from car detection to other challenging automotive tasks, such as traffic sign identification, as well as other topics in robotics and machine vision. The ability to build realistic models of different environments - such as hospital corridors or factory floors - is a challenge but within reach. Our results suggest that building synthetic images of these environments using computer graphics will be beneficial for developing new camera designs.

{\small
\bibliographystyle{ieee}
\bibliography{access}
}

\EOD

\end{document}